\documentclass[lettersize,journal]{IEEEtran}
\usepackage{amsmath,amsfonts}
\usepackage{algorithmic}
\usepackage{algorithm}
\usepackage{array}
\usepackage[caption=false,font=normalsize,labelfont=sf,textfont=sf]{subfig}
\usepackage{textcomp}
\usepackage{stfloats}
\usepackage[hyphens]{url}
\usepackage{verbatim}
\usepackage{graphicx}
\usepackage{multirow}
\usepackage{cite}
\usepackage{orcidlink}
\usepackage{tabularx}
\usepackage{booktabs}
\usepackage{rotating, arydshln}
\usepackage{threeparttablex}
\newcommand{\mathbbm}[1]{\mathbf{#1}}
\hypersetup{hidelinks}
\begin{document}

\title{SLUM-i: Semi-supervised Learning for Urban Mapping of Informal Settlements and Data Quality Benchmarking}

\author{Muhammad~Taha~Mukhtar\orcidlink{0009-0009-0833-2468}, Syed~Musa~Ali~Kazmi\orcidlink{0009-0002-9329-8039}, Khola~Naseem\orcidlink{0000-0003-4785-2588}, Muhammad~Ali~Chattha\orcidlink{0000-0002-3336-5677},
 Andreas~Dengel\orcidlink{0000-0002-6100-8255}, Sheraz~Ahmed\orcidlink{0000-0002-4239-6520}, Muhammad~Naseer~Bajwa\orcidlink{0000-0002-4821-1056}, and Muhammad~Imran~Malik\orcidlink{0000-0002-8079-5119}
\thanks{This work was supported by the German Academic Exchange Service (DAAD) under Project No.\ 57708351, titled SLUMi. 
The research was carried out at the German Research Center for Artificial Intelligence (DFKI GmbH, Kaiserslautern, Germany) and the National University of Sciences and Technology (NUST), Islamabad, Pakistan.}%
\thanks{M. T. Mukhtar, S. M. A. Kazmi, M. N. Bajwa, and M. I. Malik are with the National University of Sciences and Technology (NUST), Islamabad, Pakistan.}%
\thanks{K. Naseem, M. A. Chattha, A. Dengel, and S. Ahmed are with the German Research Center for Artificial Intelligence (DFKI), Kaiserslautern, Germany.}%
\thanks{Correspondence to: M. I. Malik (e-mail: malik.imran@seecs.edu.pk).}}%



\maketitle

\begin{abstract}
Very-high-resolution remote-sensing imagery provides a scalable basis for delineating informal settlements, but sparse annotations, severe imbalance between informal-settlement and background pixels, and cross-city heterogeneity in urban morphology and image--mask correspondence complicate model development. We present SLUM-i, a semi-supervised semantic segmentation framework together with a geographically diverse seven-city Earth observation benchmark spanning three continents. The benchmark combines a newly annotated Lahore dataset and companion Karachi and Mumbai datasets with four publicly released city datasets, totaling 14,458 RGB image--mask tiles. We quantify cross-city heterogeneity using class composition, boundary morphology, grayscale separability, correspondence between mask boundaries and image edges, and divergence between measured feature distributions. For label-efficient mapping, SLUM-i combines representation-guided unlabeled-pool curation, using embeddings from a vision foundation model (DINOv2-Small) to remove the least-similar tiles, and Class-Aware Adaptive Thresholding, which adapts pseudo-label acceptance by class through global mean-confidence and per-class mean-softmax exponential moving averages. Experiments at 10\%, 20\%, and 30\% label budgets, using convolutional and vision-transformer backbones and five random seeds, demonstrate improvements over the corresponding UniMatch baselines in multiple city--budget settings, reaching +5.9 percentage points in mean intersection-over-union. At the 30\% budget, the ResNet-101 configuration matches or exceeds its corresponding fully labeled supervised baseline in four of seven cities. Both components operate only during training and add no inference overhead.

\end{abstract}


\section{Introduction}
The United Nations estimates that $57.8\%$ of the global population lived in urban areas in 2025, with this share projected to reach $67.3\%$ by 2050~\cite{UNUrbanization2025}. In low- and middle-income countries, rapid growth can outpace formal housing markets and urban infrastructure, leading many residents to settle in informal neighborhoods. UN-Habitat defines informal settlements through three main criteria: insecure tenure, inadequate access to basic services and infrastructure, and housing that may not comply with planning or building regulations and is often located in hazardous areas~\cite{UNHabitat2015Informal}.

In Pakistan, urban slums remain largely unaccounted for in official datasets. A 2020 UNICEF survey across 10 major cities found that $53\%$ of children in slum and underserved areas were fully immunized, $56\%$ of surveyed families lived in housing built wholly or partly from non-durable materials, and $70\%$ of mothers were either illiterate or educated only through grade 5~\cite{UNICEF2020SlumCoverage}. These conditions underscore the need for targeted interventions and robust mapping, as incomplete spatial records hinder effective policy and service delivery.

Traditional surveys often fail to capture the full extent of informal settlements. Remote sensing offers a scalable alternative, and semantic segmentation enables pixel-level delineation from satellite imagery. Earlier image-based approaches used built-up indices such as NDBI and UI~\cite{Zha2003NDBI,Kawamura1997UI}, texture and vegetation information~\cite{Kuffer2016JSTARS}, and classifiers such as SVM and Random Forest~\cite{ML,ML2}, while deep fully convolutional and transformer-based methods enabled fine-scale informal-settlement mapping from very-high-resolution imagery~\cite{Persello2017GRSL,Fan2022TGRS}. Deep segmentation models, however, still require large, high-quality annotated datasets that Pakistani cities lack. Although Sentinel-2 provides additional spectral bands, its 10-m resolution limits fine-grained discrimination of formal and informal urban landscape~\cite{Gxokwe2024RSASE}. We therefore use high-resolution RGB imagery.

To reduce annotation requirements, semi-supervised remote-sensing segmentation has explored transformation consistency~\cite{Zhang2023JSTARSTransform}, iterative contrastive learning~\cite{Wang2022GRSLIterative}, label propagation~\cite{Ma2023TGRSLabelPropagation}, and class-adaptive pseudo-labeling~\cite{Wang2025JSTARS}. Fixed-threshold methods such as FixMatch and UniMatch can suppress minority-class pseudo-labels under severe imbalance, motivating adaptive alternatives~\cite{FixMatch,UniMatch,CLSIMB}. A related approach derives thresholds from per-batch class statistics~\cite{Wang2025JSTARS}. Unlabeled pools may also contain visually dissimilar tiles that create distribution mismatch. We address both problems by combining DINOv2-based unlabeled-pool filtering with Class-Aware Adaptive Thresholding (CAAT), which adapts pixel-level pseudo-label acceptance using accumulated global and class-wise confidence statistics. To our knowledge, no existing benchmark jointly evaluates semi-supervised methods across multiple cities under controlled label budgets with explicit data quality characterization. Our contributions are threefold: (i) a newly annotated Lahore dataset and companion Karachi and Mumbai datasets, forming a seven-city benchmark spanning three continents with a descriptive analysis of boundary morphology, boundary-to-image-edge alignment, measured feature distributions, and class imbalance; (ii) a framework combining DINOv2-based unlabeled-pool filtering with CAAT, evaluated with convolutional and transformer backbones; and (iii) a comprehensive evaluation against three semi-supervised and two supervised baselines.\footnote{Code, preprocessing and analysis scripts, the main per-seed results, and dataset splits are publicly available at \url{https://github.com/tahamukhtar20/Slum-i}~\cite{Mukhtar2026SLUMiZenodo}.}

\section{Data Acquisition and Processing}
\label{sec:dataset}
\subsection{Dataset Construction and Characterization}
Given the susceptibility of major urban centers to overcrowding and informal settlement expansion, this study prioritizes three of the most populous metropolitan areas in South Asia: Karachi and Lahore in Pakistan, and Mumbai in India. Data construction focuses on compiling reference polygons for informal settlements and pairing them with very-high-resolution RGB imagery.

For Lahore, we collaborated with the Katchi Abadis (Informal Settlements) Directorate of Lahore to secure the official administrative registry. Utilizing this registry, two independent annotators manually delineated $266$ distinct settlement polygons using high-resolution imagery from Google Earth~\cite{GoogleEarthPro2025} for spatial verification. Inter-annotator discrepancies were systematically resolved through a joint review analyzing local contextual indicators and official municipal records. For Karachi and Mumbai, reference datasets were derived from high-resolution satellite imagery and publicly accessible geographic annotations~\cite{MumbaiClusterMap, KarachiCartography}.

For the downstream segmentation task, a custom clustered-tiling pipeline~\cite{Mukhtar2026MapCV} transforms the WGS-84 polygons to Web Mercator (EPSG:3857), acquires Esri World Imagery~\cite{EsriWorldImagery} at zoom level 18, rasterizes polygons using pixel-center inclusion, and exports non-overlapping $512\times512$ pixel RGB image--mask pairs. The nominal ground sampling across the three city latitudes is approximately $0.51$--$0.56$\,m\,px$^{-1}$, corresponding to approximately $261$--$289$\,m per patch side. These values describe the output grid, not the native resolution of every source image in the spatially and temporally varying Esri mosaic.

To broaden evaluation across heterogeneous urban contexts, we additionally use DigitalGlobe VHR RGB imagery and reference masks from prior work~\cite{GramHansenDataset}: $0.30$\,m imagery for N.~Nairobi and $0.40$\,m imagery for El~Daein, El~Geneina, and Medell\'{i}n, all cropped to $512\times512$ pixels. The released GeoTIFFs use the respective local UTM zones; their acquisition years are not documented. The source marks the Medell\'{i}n reference annotation as less than $75\%$ complete. Across all seven datasets, ``background'' therefore means outside the available reference polygons, not independently verified formal settlement, and the masks encode mapped settlement extent rather than direct observation of tenure or service deprivation.

The tile-label distributions of the resulting datasets are detailed in Table~\ref{tab:benchmark-composition}. The benchmark exhibits a heterogeneous tile profile and strong class imbalance. A majority of the tiles in most domains contain only pixels outside the available reference polygons.

\begin{table}
    \centering
    \footnotesize
    \setlength{\tabcolsep}{3pt}
    \caption{Tile composition and class distribution of the datasets derived from per-tile raster masks. ``Mixed'' tiles denote sub-grids containing both background and informal settlement pixels; ``Slum'' tiles represent fully homogeneous informal settlement regions; ``Non-empty'' reports the proportion of total tiles containing at least one target pixel}
    \begin{tabular}{@{}lrrrrr@{}}
        \toprule
        \textbf{City} & \textbf{Total} & \textbf{Background} & \textbf{Mixed} & \textbf{Slum} & \textbf{Non-empty (\%)} \\
        \midrule
        Lahore       & 1,687 & 1,108 &   554   &  25  & 34.3\% \\
        Karachi      & 4,363 & 3,624 &   603   & 136  & 16.9\% \\
        Mumbai       & 3,741 & 2,405 & 1,336   &   0  & 35.7\% \\
        El Daein     & 2,585 & 1,508 &   555   & 522  & 41.7\% \\
        El Geneina   & 1,936 & 1,208 &   535   & 193  & 37.6\% \\
        N. Nairobi   &   111 &    58 &    53   &   0  & 47.7\% \\
        Medell\'{i}n &    35 &     8 &    24   &   3  & 77.1\% \\
        \bottomrule
    \end{tabular}
    \label{tab:benchmark-composition}
\end{table}

\begin{figure*}[t]
    \centering
    \includegraphics[width=0.96\textwidth]{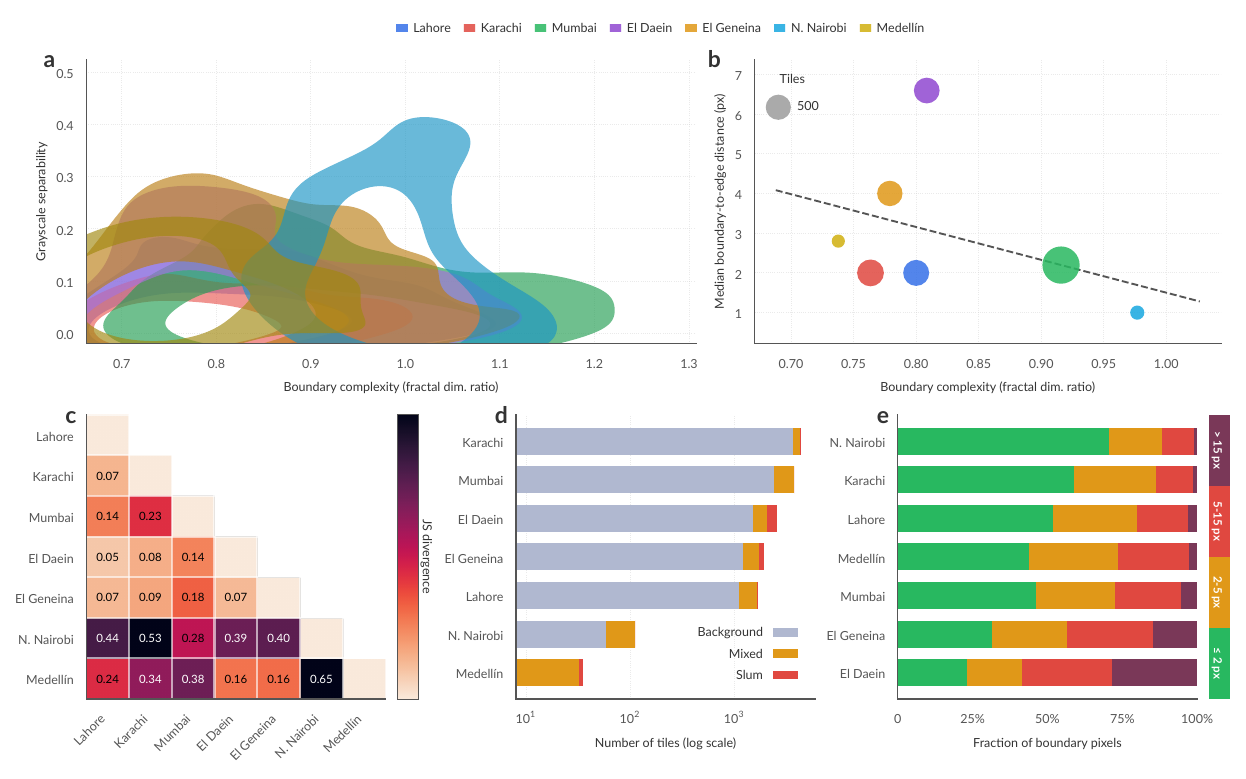}
    \caption{Descriptive dataset characterization. (a) Kernel density estimates of grayscale separability versus boundary complexity. (b) City medians of per-tile boundary complexity and median mask-boundary-to-image-edge distance; bubble area encodes the number of mixed tiles, and the dashed line is a descriptive least-squares fit. (c) Pairwise Jensen--Shannon (JS) divergence between the joint boundary-complexity and grayscale-separability distributions. (d) Log-scaled tile composition. (e) Equal-tile-weighted mean fractions of mask-boundary pixels in four distance bands. Panels (b) and (e) use only mixed tiles.}
    \label{fig:data_quality}
\end{figure*}

\subsubsection{Dataset Complexity Analysis}
\label{sec:dataset-complexity}

For each mixed tile $t$, let $I_t$ denote its 8-bit grayscale image,
$M_t\in\{0,1\}^{512\times512}$ its mask, and $B_t$ the one-pixel external
contour of $M_t$. Let
$E_t^{a,b}=\operatorname{Canny}(G_{5\times5}*I_t;a,b)$, where $G_{5\times5}$
is Gaussian smoothing. For a binary point set $Z$, $N_Z(s)$ counts the grid
cells of side $s\in\mathcal{S}=\{2^9,2^8,\ldots,2^2\}$ that intersect $Z$.
The tile-characterization variables are
\begin{equation}
\begin{aligned}
(\hat\alpha_Z,\hat\beta_Z)
 &=\arg\min_{\alpha,\beta}\sum_{s\in\mathcal S}
 [\log\!\max\{N_Z(s),1\}-\alpha-\beta\log s]^2,\\
D(Z)&=-\hat\beta_Z,\qquad
C_t=\frac{D(B_t)}{D(M_t\cap E_t^{100,200})},\\
S_t&=\frac{(\mu_{t,1}-\mu_{t,0})^2}
{\sigma_{t,1}^2+\sigma_{t,0}^2+10^{-6}},
\end{aligned}
\label{eq:tile-characterization}
\end{equation}
where $C_t$ is the boundary-to-internal-edge box-counting ratio and $S_t$ is
a Fisher-like grayscale contrast; $\mu_{t,k}$ and $\sigma_{t,k}^2$ are the
population mean and variance for pixels with $M_t=k$. Following the released
implementation, $D(Z)=0$ for fewer than 20 points, $C_t$ is omitted when its
denominator is at most $0.1$, and $S_t$ is omitted when either class contains
fewer than 50 pixels.

Image--mask boundary correspondence is summarized for mixed tiles by
\begin{equation}
\begin{aligned}
d_t(p)&=\min_{q\in E_t^{50,150}}\lVert p-q\rVert_2,
       &&p\in B_t,\\
m_t&=\operatorname*{median}_{p\in B_t}d_t(p),\\
F_t(\tau)&=\frac{1}{|B_t|}\sum_{p\in B_t}
\mathbbm{1}[d_t(p)\leq\tau],
       &&\tau\in\{2,5,15\}.
\end{aligned}
\label{eq:boundary-correspondence}
\end{equation}
The resulting distances are approximate and are reported in processed-image
pixels. Panel~(b) reports city
medians of $C_t$ and $m_t$, while panel~(e) reports city means of
$F_t(\tau)$ with equal weight per mixed tile.

For each city pair, panel~(c) bins valid $(C_t,S_t)$ samples on a common
pair-specific $15\times15$ grid spanning the pooled 1st--99th percentiles.
After adding $10^{-12}$ to each cell and normalizing the histograms $P$ and
$Q$, the reported divergence is
\begin{equation}
\begin{aligned}
\operatorname{JS}(P,Q)
 &=\tfrac12\operatorname{KL}(P\Vert A)
 +\tfrac12\operatorname{KL}(Q\Vert A),\\
A&=\tfrac12(P+Q).
\end{aligned}
\label{eq:js-characterization}
\end{equation}
using natural logarithms~\cite{JSD}. These diagnostics describe morphology,
grayscale contrast, and boundary correspondence at the processed-image scale;
they are neither sensor-invariant radiometric quantities nor direct measures
of socioeconomic conditions or annotation error.

The cities span different regimes of these model-input characteristics, as
summarised in Fig.~\ref{fig:data_quality}. The joint kernel density estimate
of $C_t$ and $S_t$ (\textit{a}) reveals two qualitatively distinct groups.
Mumbai and N.~Nairobi are structurally the most
complex cities. Mumbai exhibits a broad complexity distribution, with a median ratio
of 0.916 and a 90th percentile of approximately 1.21, consistent with varied and
intricate mapped boundaries. N.~Nairobi forms
a compact but high-separability cluster centred near fractal dimension 1.0, showing
strong normalized grayscale contrast under this diagnostic. The remaining
cities concentrate at lower complexity ($\leq 0.85$), yet show considerable within-city variance and
substantial inter-city overlap, confirming that no single complexity regime
characterises every dataset.

Boundary-to-image-edge distance provides an image-based proxy for mask alignment
(\textit{b}). It measures the distance from each mask-boundary pixel to the nearest
Canny image edge and should not be interpreted as a direct measurement of annotation
error. N.~Nairobi has the smallest median distance ($1.0$\,px; a mean per-tile
$70.5\%$ of boundary pixels within $2$\,px), while El~Daein has the largest
median distance ($7.0$\,px; $23.1\%$ within $2$\,px). Panel~(\textit{e}) further shows
that the Sudanese cities, El~Daein and El~Geneina, have the largest fractions of
mask-boundary pixels more than 15\,px from a detected image edge.

The pairwise Jensen--Shannon divergences (\textit{c}) compare each city's joint
distribution of boundary-complexity ratio and grayscale separability. Divergence is
low for Lahore--Karachi ($\mathrm{JS}{=}0.07$) and El~Daein--El~Geneina
($\mathrm{JS}{=}0.07$), whereas N.~Nairobi lies at least $\mathrm{JS}{=}0.28$ from
every other city and the Medell\'{i}n--N.~Nairobi pair reaches
$\mathrm{JS}{=}0.65$. These values characterize shift in the two measured features,
not general image-domain shift.

\begin{figure}[b]
    \centering
    \begin{minipage}{0.8\linewidth}
        \centering
        \begin{minipage}[t]{0.485\linewidth}
            \centering
            \includegraphics[width=\linewidth]{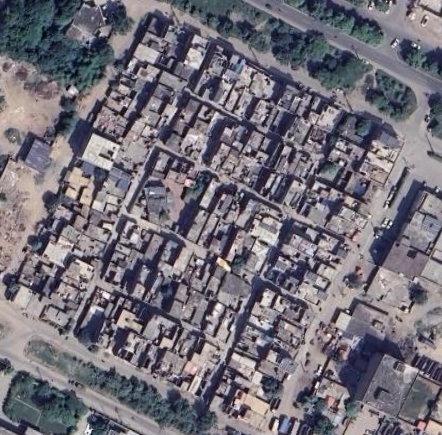}
        \end{minipage}%
        \hspace{0.01\linewidth}%
        \begin{minipage}[t]{0.485\linewidth}
            \centering
            \includegraphics[width=\linewidth]{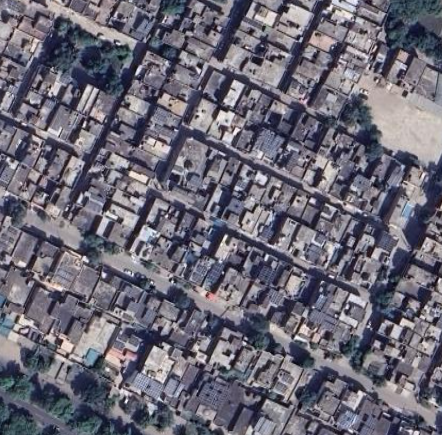}
        \end{minipage}
    \end{minipage}
    \caption{Visual contrast between informal and formal urban fabric in Lahore. Left: Zia Colony, a verified informal settlement~\cite{ZiaColony}; right: Township, a planned development. Both scenes contain dense built-up texture, showing why spectral appearance alone is insufficient for reliable slum segmentation. Source:~\cite{GoogleEarthPro2025}}
    \label{fig:distinct}
\end{figure}

Scale and class imbalance introduce a final axis of difficulty (\textit{d}). The overall 
volume of data varies dramatically, spanning multiple orders of magnitude across the cities.
Furthermore, the internal label distributions are heavily skewed; background and mixed tiles
dominate, while purely homogeneous settlement tiles are exceedingly rare.

Fig.~\ref{fig:distinct} highlights the visual similarity between the target and non-target classes. While standard land-cover segmentation often relies on clear visual contrasts (e.g., built-up areas versus vegetation), identifying informal settlements requires distinguishing between urban environments that can look structurally alike.

\subsubsection{Data Partitioning}
Given our semi-supervised learning strategy, we define three scarcity protocols
by withholding ground-truth annotations within the training set,
as detailed in Table \ref{tab:ssl_protocols}. We simulate scenarios with 10\%, 20\%,
and 30\% labeled data, treating the remaining portions as unlabeled. We employ a nested sampling strategy, i.e., the 10\% labeled set is
a strict subset of the 20\% set, which is in turn a strict subset of the
30\% set. This reduces variation between label-budget protocols because each larger labeled set adds samples to the preceding one rather than drawing an independent subset. All data splits are provided in the project repository.

\begin{table}[!h]
    \centering
    \caption{SSL label budget protocols. $D_L$ and $D_U$ denote the labeled
    and unlabeled partitions of the training set, with $D_L^{10\%} \subset
    D_L^{20\%} \subset D_L^{30\%}$}
    \begin{tabularx}{\linewidth}{Xrr}
        \toprule
        \textbf{Protocol} & \textbf{Labeled ($D_L$)} & \textbf{Unlabeled ($D_U$)} \\
        \midrule
        10\% Label & 10\% & 90\% \\
        20\% Label & 20\% & 80\% \\
        30\% Label & 30\% & 70\% \\
        \bottomrule
    \end{tabularx}
    \label{tab:ssl_protocols}
\end{table}

\section{Methodology}
\label{sec:methodology}
\subsection{Semi-supervised Learning}
The proposed framework extends the UniMatch pipeline~\cite{UniMatch}, implemented here with a ResNet-101 backbone and a DeepLabV3+~\cite{DeepLab} segmentation head. In highly imbalanced informal-settlement maps, UniMatch's static confidence threshold ($\tau = 0.95$) can reject lower-confidence minority-class pseudo-labels. In our experiments, this fixed-threshold pipeline sometimes falls below the corresponding supervised baseline.

\begin{figure*}[t]
    \centering
    \includegraphics[width=1\linewidth]{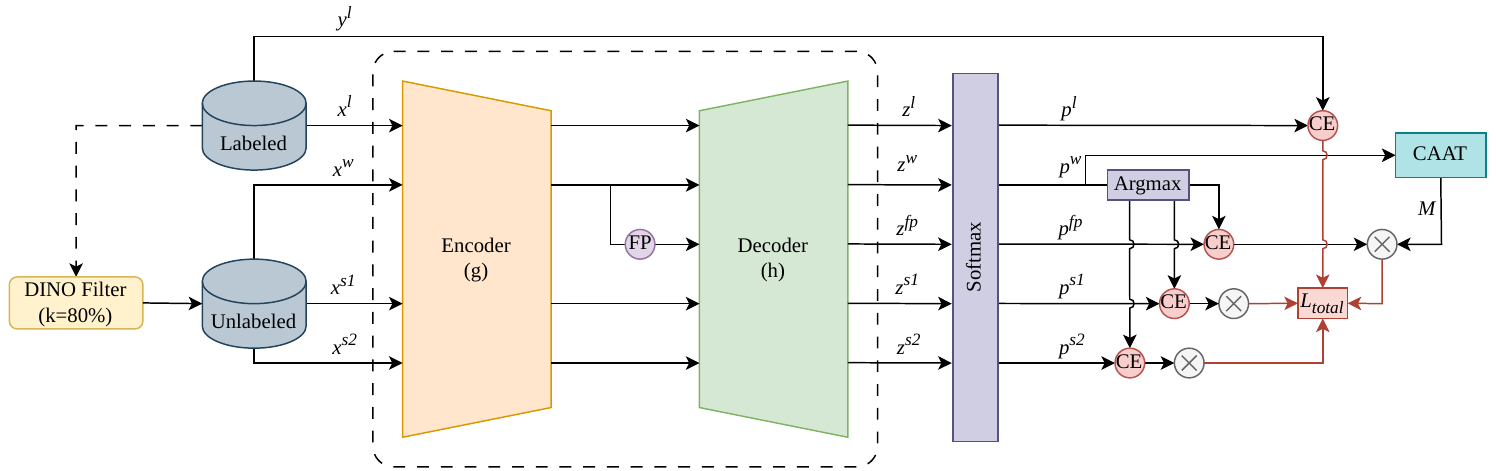}
    \caption{Overview of the proposed semi-supervised framework. The pipeline consists of a shared encoder-decoder network processing labeled ($x^l$), weak ($x^w$), and strong ($x^{s1}, x^{s2}$) views. Prior to training, the unlabeled pool is curated by a DINO-based filter that retains only tiles visually similar to the labeled set. During training, the Class-Aware Adaptive Threshold (CAAT) dynamically gates pseudo-label contributions via the binary mask $M$.}
    \label{fig:architecture}
\end{figure*}

To address these limitations, we introduce two modular components designed to stabilize training and minimise degradation in slum detection settings, as illustrated in Fig.~\ref{fig:architecture}:
\begin{enumerate}
    \item A DINO-embedding-based unlabeled-pool filter that excludes low-similarity tiles before training as a proxy for reducing mismatch between the labeled and unlabeled data streams.
    \item A Class-Aware Adaptive Threshold (CAAT) that replaces the static confidence requirement with a per-class threshold tracked via exponential moving averages (EMAs), reducing minority-class suppression.
\end{enumerate}

\subsubsection{Baseline Framework}
Our method is inspired by UniMatch~\cite{UniMatch}, which adapts the FixMatch \cite{FixMatch} weak-to-strong consistency framework. The training process operates on a single network architecture consisting of an encoder $g(\cdot)$ and a decoder $h(\cdot)$, processing labeled and unlabeled data streams. For an input $x$, we denote the intermediate feature embedding as $v = g(x)$ and the final output logits as $z = h(v)$. The softmax probability is $p = \operatorname{softmax}(z)$.

For the supervised component, a labeled image $x^l$ is passed through the model to generate logits $z^l$, which are processed via softmax to produce $p^l$. The supervised loss $\mathcal{L}_{sup}$ is then calculated as the standard cross-entropy between $p^l$ and the ground-truth label $y^l$.

The unsupervised component enforces consistency across different perturbations of the same input. An unlabeled image $x^u$ first undergoes weak augmentation to obtain $x^w$. The model predicts logits $z^w = h(g(x^w))$, where the softmax probability $p^w = \operatorname{softmax}(z^w)$ serves as the source for pseudo-label generation. To enforce consistency, the same image simultaneously undergoes strong augmentation (specifically CutMix~\cite{CutMix}) to generate two views, $x^{s1}$ and $x^{s2}$. The model is then optimized to align the predictions from these strongly distorted inputs with the pseudo-labels derived from the weak view $x^w$. Additionally, a Feature Perturbation (FP) mechanism injects noise into the encoder features of the weak view to produce $z^{fp}$, further enforcing internal feature robustness. The corresponding softmax probabilities for the strong and perturbed views are denoted as $p^{s1}$, $p^{s2}$, and $p^{fp}$.

\subsubsection{DINO-based Unlabeled Pool Filtering}
\label{sec:dino-filter}
A core challenge in applying SSL to slum mapping is that the unlabeled pool can contain visually dissimilar tiles (e.g., dense vegetation, highways, or highly structured formal housing) that differ substantially from the labeled distribution. Forcing the model to generate pseudo-labels on these regions can introduce noise. To address this, we curate the unlabeled pool prior to training using similarity in a frozen DINOv2~\cite{Dinov2} representation space.

For a given city, let $D_L$ and $D_U$ denote the labeled and unlabeled tile sets, respectively. We extract an L2-normalized CLS token embedding from a frozen DINOv2-Small backbone for each tile, denoted as $f(\cdot) \in \mathbb{R}^d$ (where $d=384$), and define the mean labeled embedding as $\mathbf{c}_L = \frac{1}{|D_L|} \sum_{x^l \in D_L} f(x^l)$. The released implementation assigns each unlabeled tile $x^u \in D_U$ its mean cosine similarity to all labeled tiles:
\begin{equation}
    s(x^u) = \frac{1}{|D_L|}\sum_{x^l \in D_L} f(x^u)^\top f(x^l)
    = f(x^u)^\top \mathbf{c}_L.
\end{equation}
Because $\|\mathbf{c}_L\|$ is constant within each city and label budget, this score gives the same ranking as cosine similarity to the labeled centroid.
We sort the unlabeled pool by $s(x^u)$ and retain the top $k=80\%$ of tiles, excluding the bottom $20\%$ as low-similarity samples from SSL training.

This filtering procedure is motivated by importance weighting and covariate shift theory~\cite{Shimodaira2000}. Semi-supervised generalization bounds typically assume that the marginal distributions of the labeled and unlabeled data match ($p(x^l) \approx p(x^u)$). Filtering via self-supervised embedding similarity is intended to reduce this mismatch at the image level before training begins.

\subsubsection{Class-Aware Adaptive Thresholding (CAAT)}
Standard UniMatch relies on a fixed global threshold ($\tau=0.95$), which can disproportionately suppress the minority class under severe imbalance. We adapt FreeMatch-style self-adaptive thresholding~\cite{FreeMatch} to pixel-wise segmentation and denote the resulting mechanism CAAT. At each iteration $t$, it updates two EMA quantities over valid (non-ignored) pixels $\mathcal{V}$:
\begin{equation}
    \tilde{p}^{(t)} = \beta\,\tilde{p}^{(t-1)} + (1-\beta)\,\frac{1}{|\mathcal{V}|}\sum_{(i,j)\in\mathcal{V}} \max_c\, p^w_{i,j,c},
\end{equation}
\begin{equation}
    \mu^{(t)} = \beta\,\mu^{(t-1)} + (1-\beta)\,\frac{1}{|\mathcal{V}|}\sum_{(i,j)\in\mathcal{V}} p^w_{i,j},
\end{equation}
where $\beta=0.999$ controls EMA momentum, $\tilde{p}^{(t)}$ is a global mean-confidence EMA, and $\mu^{(t)} \in \mathbb{R}^C$ is the per-class mean-softmax EMA (where $C$ denotes the total number of target classes), initialised uniformly to $1/C$. A normalised class modulator is then computed as:
\begin{equation}
    \phi_c^{(t)} = \frac{\mu_c^{(t)}}{\max_{c'} \mu_{c'}^{(t)}}.
\end{equation}
The per-pixel adaptive threshold for a pixel with pseudo-label prediction $\hat{y}_{i,j} = \arg\max_c p^w_{i,j,c}$ is:
\begin{equation}
    \tau_{i,j}^{(t)} = \min\!\left(\tilde{p}^{(t)} \cdot \phi_{\hat{y}_{i,j}}^{(t)},\; 0.95\right).
\end{equation}
This scales the global threshold down for underrepresented classes (low $\phi_c$), while the dominant class remains at the global threshold, with an upper bound of $0.95$ following standard SSL protocols~\cite{FixMatch, UniMatch, UniMatchv2}. Finally, the binary mask that gates pixel contributions to the unsupervised loss is:
\begin{equation}
    M_{i,j} = \mathbbm{1}\!\left[\, \max_c\, p^w_{i,j,c} \;\geq\; \tau_{i,j}^{(t)} \,\right].
\end{equation}
This establishes a dynamic curriculum; underrepresented slum pixels receive lower thresholds early in training, preventing them from being systematically discarded as the fixed-threshold baseline would.

\subsubsection{Total Objective Function}
In the ResNet-101 implementation, the unsupervised objective combines the Strong Augmentation loss ($\mathcal{L}_{s}$) and the Feature Perturbation loss ($\mathcal{L}_{fp}$). For the strong views ($s \in \{s1, s2\}$), the loss is gated by the CAAT binary mask $M$, where $|\mathcal{P}|$ is the total number of valid pixels:
\begin{equation}
    \mathcal{L}_{s} = \frac{1}{|\mathcal{P}|} \sum_{(i,j) \in \mathcal{P}} M_{i,j} \cdot \ell_{ce}(p^s_{i,j}, \hat{y}_{i,j}).
\end{equation}
For the feature perturbation stream, the same adaptive threshold mask $M$ is applied:
\begin{equation}
    \mathcal{L}_{fp} = \frac{1}{|\mathcal{P}|} \sum_{(i,j) \in \mathcal{P}} M_{i,j} \cdot \ell_{ce}(p^{fp}_{i,j}, \hat{y}_{i,j}).
\end{equation}
For the ResNet-101 implementation, the final training objective combines the supervised and unsupervised components using the weights in the released training code:
\begin{equation}
    \mathcal{L}_{total}^{R} = \frac{1}{2}\mathcal{L}_{sup}
    + \frac{1}{8}(\mathcal{L}_{s1} + \mathcal{L}_{s2})
    + \frac{1}{4}\mathcal{L}_{fp}.
\end{equation}
The DINOv2 implementation follows UniMatch-v2 and omits the feature-perturbation stream, giving
\begin{equation}
    \mathcal{L}_{total}^{D} = \frac{1}{2}\mathcal{L}_{sup}
    + \frac{1}{4}(\mathcal{L}_{s1} + \mathcal{L}_{s2}).
\end{equation}
Both formulations learn from confident, class-aware pseudo-labels via CAAT, while the ResNet-101 formulation additionally maintains feature robustness through perturbation.

\subsection{Experimental Setup}
The proposed framework was implemented in PyTorch and trained on DFKI's Pegasus Compute Cluster (B200/RTXB6000/H200/H100/A100/RTXA6000/L40S). We maintain a consistent data augmentation pipeline, including random horizontal flipping and scaling ($0.8$ to $1.2\times$). All experiments are repeated over five random seeds ($\{0, 42, 123, 999, 1337\}$), and results are reported as mean $\pm$ population standard deviation across the five runs.

Our primary experiments utilize a ResNet-101~\cite{ResNet} backbone with a DeepLabV3+~\cite{DeepLab} head. We train for $80$ epochs with a batch size of $8$ and a crop size of $512 \times 512$. We employ the SGD optimizer with a momentum of $0.9$, a weight decay of $1 \times 10^{-4}$, and an initial learning rate ($\eta$) of $0.02$.

To evaluate the modularity of our approach within the current paradigm of foundation models and self-supervised representation learning, we integrated the combined framework into the UniMatch-v2~\cite{UniMatchv2} pipeline. This configuration employs a DINOv2-Small~\cite{Dinov2} backbone coupled with a DPT~\cite{DPT} decoder. Following the standard UniMatch-v2 protocol, we use a crop size of $518 \times 518$ to satisfy the $14 \times 14$ patch-size alignment required by the Vision Transformer (ViT)~\cite{ViT} architecture. The backbone is frozen, focusing learning on the segmentation head. Training is conducted for $60$ epochs using the AdamW optimizer ($\beta_1=0.9, \beta_2=0.999$) with a weight decay of $0.01$, a base learning rate of $5 \times 10^{-6}$, and a $40\times$ multiplier. Pseudo-labels are generated by an EMA teacher whose decay at iteration $t$ is $\alpha_t=\min(1-1/(t+1),0.996)$. Within each backbone family, the baseline and proposed variants use the same optimization schedule and augmentation pipeline; the proposed variants add the unlabeled-pool filter and adaptive thresholding described above.

\section{Results}
\label{sec:experimentation}
\subsection{Quantitative Evaluation}
\begin{table*}[!t]
    \centering
    \small
    \setlength{\tabcolsep}{1.5pt}
    \renewcommand{\arraystretch}{1.08}
    \caption{Mean intersection-over-union (mIoU) by city and label budget, reported as mean $\pm$ population standard deviation over five seeds. Within each label-budget and backbone block, $\mathbf{bold}$ denotes the best result and \underline{underline} the second best; ResNet-101 and DINOv2 are ranked separately. Rankings are based on unrounded means. The $100\%$ rows are unranked supervised reference baselines trained without unlabeled data. A dagger ($^{\dagger}$) denotes methods using a DINOv2~\cite{Dinov2} backbone}
    \begin{tabularx}{\textwidth}{l@{\hspace{5pt}}l *{7}{>{\centering\arraybackslash}X}}
        \toprule
        \textbf{Budget} & \textbf{Method} & \textbf{El Daein} & \textbf{El Geneina} & \textbf{N. Nairobi} & \textbf{Medell\'{i}n} & \textbf{Mumbai} & \textbf{Lahore} & \textbf{Karachi} \\
        \midrule
        \multirow{7}{*}{\textbf{10\%}} & Supervised & $0.654$$\pm$0.018 & $0.629$$\pm$0.035 & $0.662$$\pm$0.005 & $0.230$$\pm$0.098 & $0.586$$\pm$0.008 & $\mathbf{0.507}$$\pm$0.013 & $0.496$$\pm$0.016 \\
         & FixMatch~\cite{FixMatch} & \underline{$0.690$}$\pm$0.021 & $0.645$$\pm$0.017 & $0.653$$\pm$0.023 & $0.782$$\pm$0.063 & $\mathbf{0.596}$$\pm$0.008 & $0.499$$\pm$0.016 & \underline{$0.516$}$\pm$0.012 \\
         & UniMatch~\cite{UniMatch} & $0.687$$\pm$0.016 & \underline{$0.667$}$\pm$0.012 & $\mathbf{0.704}$$\pm$0.010 & \underline{$0.794$}$\pm$0.076 & \underline{$0.594$}$\pm$0.008 & \underline{$0.505$}$\pm$0.012 & $\mathbf{0.530}$$\pm$0.007 \\
         & \textbf{Ours} & $\mathbf{0.723}$$\pm$0.009 & $\mathbf{0.669}$$\pm$0.039 & \underline{$0.701$}$\pm$0.008 & $\mathbf{0.853}$$\pm$0.043 & $0.593$$\pm$0.004 & $0.503$$\pm$0.005 & $0.514$$\pm$0.012 \\
        \cmidrule(lr){2-9}
         & Supervised~$^{\dagger}$ & \underline{$0.692$}$\pm$0.004 & $0.661$$\pm$0.011 & $0.687$$\pm$0.005 & $0.272$$\pm$0.083 & $0.605$$\pm$0.010 & $\mathbf{0.518}$$\pm$0.009 & $0.548$$\pm$0.006 \\
         & UniMatch-v2~\cite{UniMatchv2}$^{\dagger}$ & $0.684$$\pm$0.011 & $\mathbf{0.700}$$\pm$0.006 & \underline{$0.713$}$\pm$0.008 & $\mathbf{0.868}$$\pm$0.010 & \underline{$0.614$}$\pm$0.006 & \underline{$0.517$}$\pm$0.006 & \underline{$0.551$}$\pm$0.006 \\
         & \textbf{Ours}~$^{\dagger}$ & $\mathbf{0.698}$$\pm$0.009 & \underline{$0.693$}$\pm$0.012 & $\mathbf{0.714}$$\pm$0.004 & \underline{$0.866$}$\pm$0.033 & $\mathbf{0.615}$$\pm$0.004 & $0.516$$\pm$0.002 & $\mathbf{0.556}$$\pm$0.009 \\
        \midrule
        \multirow{7}{*}{\textbf{20\%}} & Supervised & $0.687$$\pm$0.017 & $0.635$$\pm$0.041 & \underline{$0.716$}$\pm$0.008 & $0.230$$\pm$0.098 & $0.584$$\pm$0.009 & $0.499$$\pm$0.012 & $0.503$$\pm$0.019 \\
         & FixMatch~\cite{FixMatch} & $\mathbf{0.734}$$\pm$0.014 & $0.683$$\pm$0.012 & $0.698$$\pm$0.015 & $0.816$$\pm$0.054 & \underline{$0.592$}$\pm$0.009 & \underline{$0.515$}$\pm$0.008 & $0.513$$\pm$0.022 \\
         & UniMatch~\cite{UniMatch} & \underline{$0.713$}$\pm$0.025 & $\mathbf{0.714}$$\pm$0.014 & $\mathbf{0.728}$$\pm$0.003 & \underline{$0.848$}$\pm$0.076 & $0.581$$\pm$0.023 & $0.514$$\pm$0.020 & \underline{$0.540$}$\pm$0.033 \\
         & \textbf{Ours} & $0.711$$\pm$0.022 & \underline{$0.709$}$\pm$0.009 & $0.714$$\pm$0.003 & $\mathbf{0.891}$$\pm$0.011 & $\mathbf{0.596}$$\pm$0.009 & $\mathbf{0.526}$$\pm$0.009 & $\mathbf{0.557}$$\pm$0.008 \\
        \cmidrule(lr){2-9}
         & Supervised~$^{\dagger}$ & $0.721$$\pm$0.007 & $0.707$$\pm$0.008 & $0.707$$\pm$0.003 & $0.272$$\pm$0.083 & $0.595$$\pm$0.006 & $\mathbf{0.528}$$\pm$0.010 & $0.547$$\pm$0.006 \\
         & UniMatch-v2~\cite{UniMatchv2}$^{\dagger}$ & \underline{$0.738$}$\pm$0.006 & \underline{$0.709$}$\pm$0.009 & $\mathbf{0.724}$$\pm$0.004 & \underline{$0.865$}$\pm$0.026 & \underline{$0.595$}$\pm$0.006 & \underline{$0.524$}$\pm$0.010 & \underline{$0.547$}$\pm$0.007 \\
         & \textbf{Ours}~$^{\dagger}$ & $\mathbf{0.745}$$\pm$0.007 & $\mathbf{0.715}$$\pm$0.007 & \underline{$0.724$}$\pm$0.008 & $\mathbf{0.887}$$\pm$0.024 & $\mathbf{0.608}$$\pm$0.004 & $0.519$$\pm$0.008 & $\mathbf{0.557}$$\pm$0.008 \\
        \midrule
        \multirow{7}{*}{\textbf{30\%}} & Supervised & $0.715$$\pm$0.010 & $0.690$$\pm$0.014 & $0.721$$\pm$0.005 & \underline{$0.892$}$\pm$0.008 & $0.581$$\pm$0.012 & $0.509$$\pm$0.013 & \underline{$0.527$}$\pm$0.007 \\
         & FixMatch~\cite{FixMatch} & $0.740$$\pm$0.013 & \underline{$0.712$}$\pm$0.018 & $0.725$$\pm$0.007 & $0.858$$\pm$0.029 & $\mathbf{0.596}$$\pm$0.020 & \underline{$0.517$}$\pm$0.019 & $0.510$$\pm$0.026 \\
         & UniMatch~\cite{UniMatch} & \underline{$0.745$}$\pm$0.017 & $0.710$$\pm$0.021 & $\mathbf{0.732}$$\pm$0.005 & $0.889$$\pm$0.015 & $0.588$$\pm$0.014 & $\mathbf{0.524}$$\pm$0.009 & $0.509$$\pm$0.022 \\
         & \textbf{Ours} & $\mathbf{0.747}$$\pm$0.008 & $\mathbf{0.718}$$\pm$0.007 & \underline{$0.727$}$\pm$0.006 & $\mathbf{0.896}$$\pm$0.012 & \underline{$0.589$}$\pm$0.018 & $0.503$$\pm$0.014 & $\mathbf{0.546}$$\pm$0.013 \\
        \cmidrule(lr){2-9}
         & Supervised~$^{\dagger}$ & $0.733$$\pm$0.007 & $0.704$$\pm$0.001 & $0.708$$\pm$0.007 & $0.902$$\pm$0.018 & $0.593$$\pm$0.008 & $\mathbf{0.539}$$\pm$0.008 & $0.555$$\pm$0.007 \\
         & UniMatch-v2~\cite{UniMatchv2}$^{\dagger}$ & \underline{$0.752$}$\pm$0.005 & \underline{$0.721$}$\pm$0.008 & $\mathbf{0.729}$$\pm$0.002 & \underline{$0.925$}$\pm$0.011 & \underline{$0.595$}$\pm$0.010 & $0.525$$\pm$0.010 & \underline{$0.558$}$\pm$0.010 \\
         & \textbf{Ours}~$^{\dagger}$ & $\mathbf{0.762}$$\pm$0.007 & $\mathbf{0.721}$$\pm$0.006 & \underline{$0.726$}$\pm$0.004 & $\mathbf{0.933}$$\pm$0.002 & $\mathbf{0.600}$$\pm$0.013 & \underline{$0.534$}$\pm$0.005 & $\mathbf{0.567}$$\pm$0.003 \\
        \midrule
        \multirow{2}{*}{\textbf{100\%}} & Supervised & $0.726$$\pm$0.022 & $0.719$$\pm$0.015 & $0.741$$\pm$0.008 & $0.881$$\pm$0.015 & $0.574$$\pm$0.014 & $0.526$$\pm$0.019 & $0.536$$\pm$0.014 \\
         & Supervised~$^{\dagger}$ & $0.799$$\pm$0.003 & $0.747$$\pm$0.005 & $0.715$$\pm$0.015 & $0.949$$\pm$0.002 & $0.594$$\pm$0.012 & $0.552$$\pm$0.013 & $0.576$$\pm$0.008 \\
        \bottomrule
    \end{tabularx}
    \label{tab:main_results}
\end{table*}

To assess the robustness of our proposed framework, we evaluate its performance across seven distinct urban environments under varying conditions of label scarcity. The comprehensive results, detailed in Table~\ref{tab:main_results}, report the mean intersection-over-union (mIoU) across three label budgets and two backbone architectures.

\paragraph{African and Latin American Cities.}
At the 10\% label budget, our method shows its largest gains in El~Daein and Medell\'{i}n, improving over UniMatch by +3.6\,pp and +5.9\,pp, respectively. At the 30\% budget, it achieves an mIoU of $0.896$ in Medell\'{i}n and $0.747$ in El~Daein, exceeding their corresponding full-label baselines ($0.881$ and $0.726$).

\paragraph{South Asian Cities.}
Performance varies by city and label budget. In Karachi, our method improves over UniMatch by +1.7\,pp and +3.7\,pp at the 20\% and 30\% budgets, respectively. In Mumbai, it achieves the best result at 20\% and remains close to the strongest baseline at 10\% and 30\%. In Lahore, our ResNet-101 configuration exceeds the corresponding supervised model at 20\%, while the supervised DINOv2 model remains strongest across the three label budgets.

\paragraph{N.~Nairobi.}
At the 10\% and 30\% label budgets, the difference between our method and UniMatch in N.~Nairobi falls within one standard deviation; at 20\%, our method trails UniMatch by 1.4\,pp.

\paragraph{DINOv2 backbone.}
Integrating the combined framework into the UniMatch-v2~\cite{UniMatchv2} pipeline with a frozen DINOv2-Small backbone yields a consistent improvement pattern. Our method achieves the best performance in 5 of 7 cities at the 20\% and 30\% label budgets, and in 4 of 7 cities at the 10\% budget. These results show that the combined framework integrates into both evaluated backbone families without structural modification.

\paragraph{Comparison with full-label baselines.}
At the 30\% label budget, our ResNet-101 model matches or exceeds the corresponding full-label ResNet-101 baseline in 4 of 7 cities (El~Daein, Medell\'{i}n, Mumbai, and Karachi), using only 30\% of the labeled data alongside unlabeled tiles. Under the DINOv2 backbone, our method at the 30\% label budget approaches the full-label baseline despite using substantially fewer annotations. In particular, the gaps are $0.762$ versus $0.799$ in El~Daein and $0.933$ versus $0.949$ in Medell\'{i}n.

\subsection{Qualitative Analysis}
\label{sec:qualitative}

\begin{figure*}[t]
\centering
\includegraphics[width=0.96\textwidth]{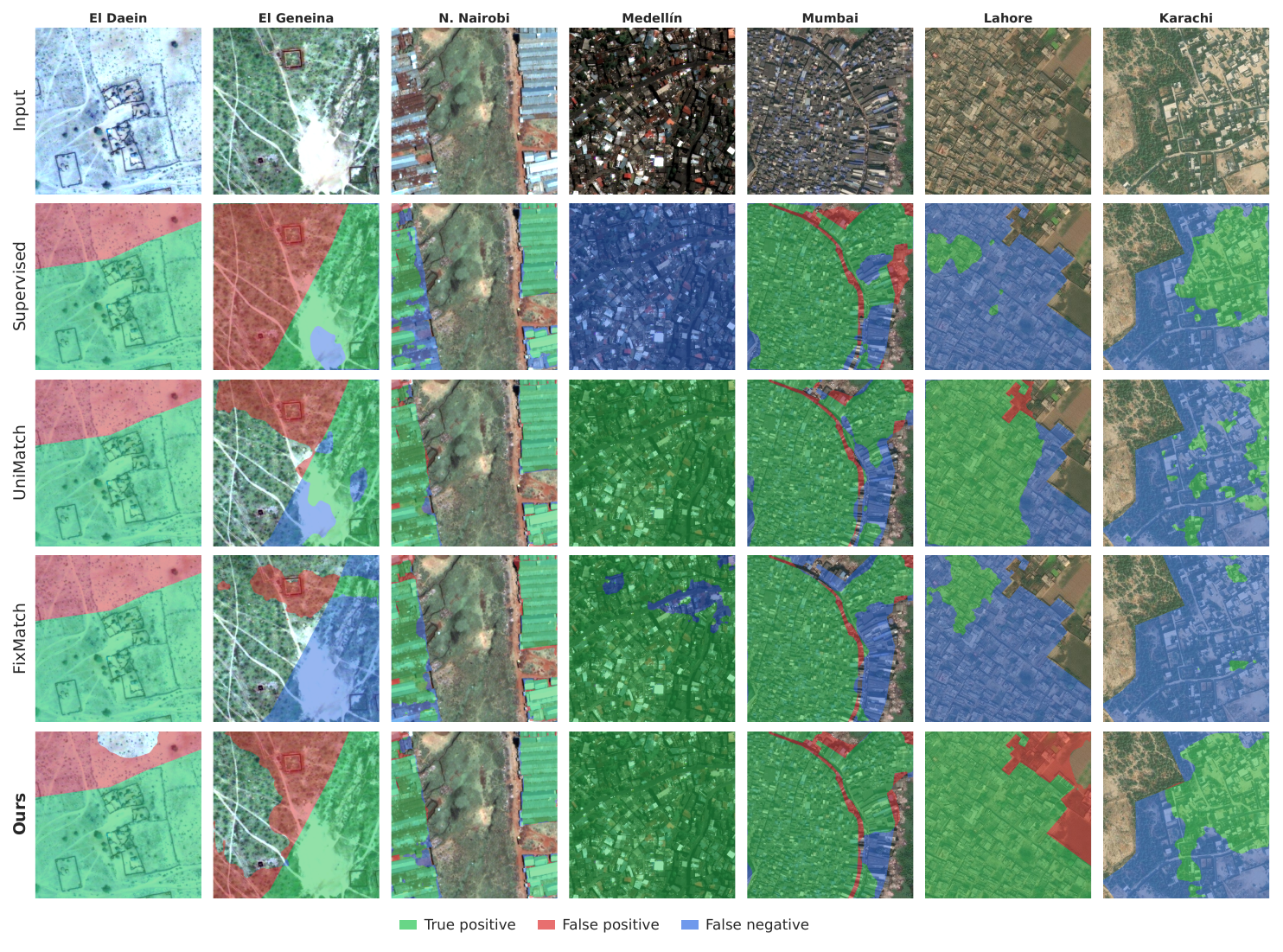}
\caption{Qualitative comparison between the baselines and our method across all datasets at the $10\%$ label budget. Imagery sources: Esri World Imagery for Lahore, Karachi, and Mumbai~\cite{EsriWorldImagery}; external-city imagery from Gram-Hansen et al.~\cite{GramHansenDataset}.}
\label{fig:qualitative}
\end{figure*}

Qualitative results support the trends observed in the quantitative evaluation. At the
$10\%$ label budget, each row in Fig.~\ref{fig:qualitative} corresponds to one method
and each column to one city.
Supervised-only predictions (row~2) reveal a characteristic failure mode under label
scarcity: large contiguous false-negative regions, where the model fails to
segment slum regions and instead defaults to the background class under severe class
imbalance. FixMatch and UniMatch partially recover these regions but also introduce
additional false positives in structurally ambiguous formal-housing areas.
Our method (bottom row) often reduces the extent of false-negative regions,
particularly in El~Daein, Medell\'{i}n, and Karachi. This behaviour is consistent with
the intended effect of the class-aware thresholding mechanism, which preserves a larger
fraction of lower-confidence slum pseudo-labels compared to the fixed-threshold
baseline ($\tau = 0.95$).

These qualitative observations align with the quantitative trends reported in
Table~\ref{tab:main_results}, where the largest numerical improvements are generally
associated with more spatially coherent prediction boundaries and fewer large omission
regions.

\subsection{Ablation Studies}

\begin{figure*}[t]
    \centering
    \includegraphics[width=0.96\textwidth]{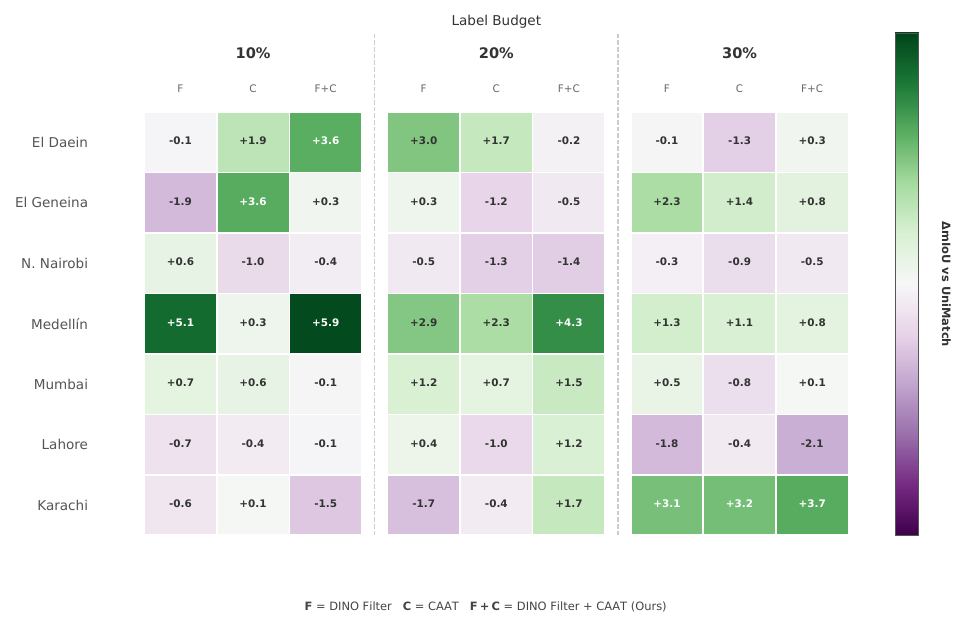}
    \caption{Component-wise ablation study of the DINO-based unlabeled pool filter (F), Class-Aware Adaptive Thresholding (C), and their combined configuration (F+C) across cities and label budgets. Each cell reports the change in mIoU, in percentage points, relative to the UniMatch baseline; positive values indicate improved performance.
    }
    \label{fig:ablation}
\end{figure*}

The ablation study isolates the contribution of each proposed component by comparing
three configurations against the UniMatch baseline: the DINO filter alone (F),
CAAT alone (C), and their combination (F+C, i.e.\ our full method). Each cell in
Fig.~\ref{fig:ablation} reports $\Delta$mIoU in percentage points.

\paragraph{DINO filter alone.}
Filtering the unlabeled pool by embedding similarity (F) produces its largest gains in Medell\'{i}n ($+5.1$\,pp at 10\% and $+2.9$\,pp at 20\%) and Karachi ($+3.1$\,pp at 30\%).

\paragraph{CAAT alone.}
CAAT alone (C) delivers its strongest gains in El~Geneina at 10\% ($+3.6$\,pp), El~Daein at 10\%--20\% ($+1.9$\,pp and $+1.7$\,pp), and Karachi at 30\% ($+3.2$\,pp).

\paragraph{Combined method.}
The full method (F+C) achieves the strongest aggregate performance across cities
and label budgets. The two components appear complementary rather than redundant:
filtering improves the quality of the unlabeled pool entering training, while
CAAT improves the utilisation of valid but lower-confidence slum pixels
within that pool.
This interaction is particularly visible in Karachi at the 20\% budget, where
neither component alone improves over UniMatch (F$=-1.7$\,pp; C$=-0.4$\,pp),
yet their combination yields a positive gain (F+C$=+1.7$\,pp).

\section{Discussion}
\subsection{Dataset Complexity and SSL Performance}

The complexity analysis in Section~\ref{sec:dataset-complexity} provides descriptive context for the results in Table~\ref{tab:main_results}. The largest gains occur in El~Daein and Medell\'{i}n at the 10\% label budget and in Karachi at the 30\% budget. The measured city attributes also vary substantially: boundary morphology and mask-boundary-to-image-edge distance differ across cities, while the Jensen--Shannon analysis quantifies differences in the joint boundary-complexity and grayscale-separability distributions~\cite{JSD}. These descriptors characterize cross-city differences rather than serve as causal predictors of SSL performance.

The proposed components address different stages of pseudo-label learning. The DINO filter removes the lowest-similarity unlabeled tiles before pseudo-label generation, while CAAT lowers the acceptance threshold when confidence for the slum class remains low. Their combined performance indicates complementary roles in unlabeled-pool curation and pseudo-label acceptance.

\subsection{Architecture Generality}

The combined framework was evaluated under both ResNet-101/DeepLabV3+ and DINOv2-Small/DPT configurations. Its improvement pattern under both configurations demonstrates compatibility with convolutional and vision-transformer backbones without requiring structural modification.

\subsection{Limitations}

The evaluation uses patch-level partitions and includes two relatively small external datasets, N.~Nairobi and Medell\'{i}n, with the source flagging the Medell\'{i}n annotation as less than $75\%$ complete. Future work will evaluate geographically held-out partitions, city-adaptive retention thresholds, joint multi-city training, and finer-grained settlement categories.

\section{Conclusion}
We introduced SLUM-i, a semi-supervised framework and multi-city benchmark for satellite-based informal-settlement segmentation. The four-dimensional analysis documents substantial variation in boundary morphology, boundary-to-image-edge alignment, measured feature distributions, and class composition across seven geographically diverse cities.

The combined DINO-based unlabeled-pool filter and Class-Aware Adaptive Threshold achieves gains of up to +5.9 percentage points in mIoU over UniMatch and adds no inference overhead. The filter removes low-similarity unlabeled tiles before training, while CAAT reduces minority-class suppression during pseudo-label selection. The combined framework is compatible with both evaluated backbone families. Under ResNet-101, it matches or exceeds the corresponding full-label baseline in four cities at the 30\% label budget while using only a fraction of the available annotations. We publicly release all dataset splits, the main per-seed results, and analysis scripts to support reproducibility and future benchmarking in this underserved domain.

\section*{Acknowledgment}
The authors would like to thank ChatGPT (developed by OpenAI) and Claude (developed by Anthropic) as assistive tools in refining and editing portions of the manuscript.

\bibliographystyle{IEEEtran}
\bibliography{bib/document}

\begin{IEEEbiography}[{\includegraphics[width=1in,height=1.25in,clip,keepaspectratio]{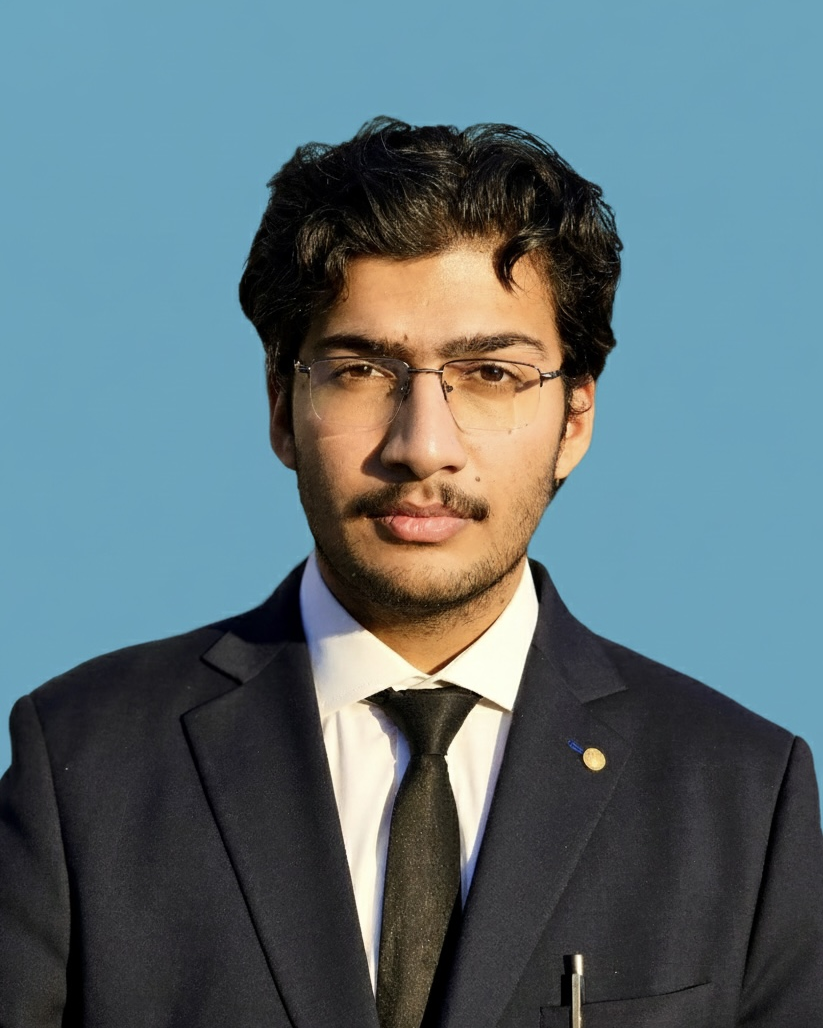}}]{Muhammad Taha Mukhtar} received the B.S. degree in computer science from the National University of Sciences and Technology (NUST), Islamabad, Pakistan, in 2025. He was a DAAD-funded Research Intern at the German Research Center for Artificial Intelligence (DFKI) and is currently a Research Assistant at the National Center of Artificial Intelligence (NCAI), Islamabad. His research focuses on remote sensing and computer vision.
\end{IEEEbiography}
\begin{IEEEbiography}[{\includegraphics[width=1in,height=1.25in,clip,keepaspectratio]{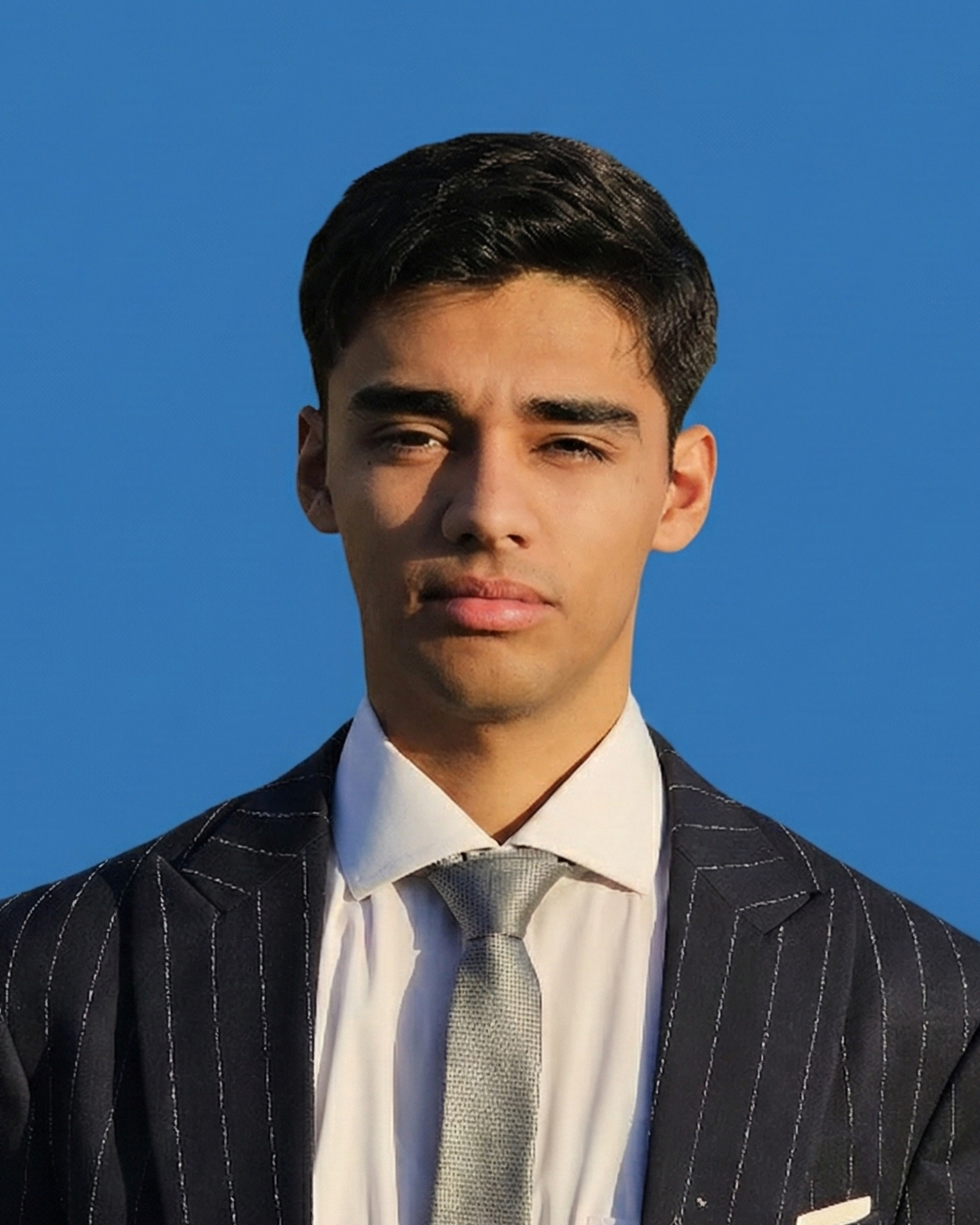}}]{Syed Musa Ali Kazmi} received the B.S. degree in computer science from the National University of Sciences and Technology (NUST), Islamabad, Pakistan, in 2025. He has gained industrial research experience at DataQuartz and EMRChains, where he contributed to the development of AI-driven solutions. His research interests include deep learning, computer vision, reinforcement learning, and large language models (LLMs).
\end{IEEEbiography}
\begin{IEEEbiography}[{\includegraphics[width=1in,height=1.25in,clip,keepaspectratio]{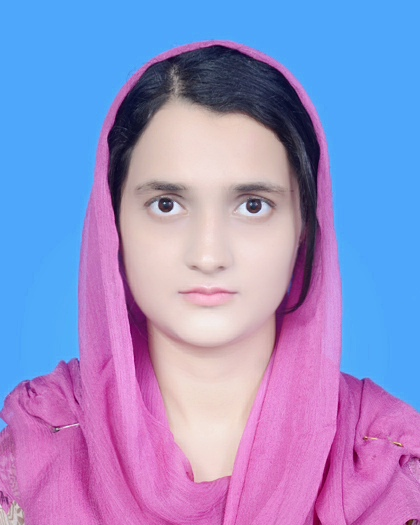}}]{Khola Naseem}
is a PhD researcher at RPTU Kaiserslautern, Germany. She received her MS in Computer Science from NUST, Islamabad, Pakistan, in 2021. Her research focuses on deep learning and computer vision, with recent work on applying deep learning to microscopic images.
\end{IEEEbiography}
\begin{IEEEbiography}[{\includegraphics[width=1in,height=1.25in,clip,keepaspectratio]{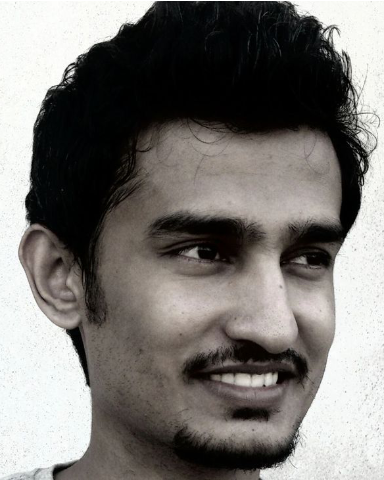}}]{Muhammad Ali Chattha} received the master's degree in electrical engineering from Lahore University of Management Sciences, Pakistan. He is pursuing the Ph.D. degree in computer science at Technische Universität Kaiserslautern with the German Research Center for Artificial Intelligence (DFKI GmbH). His research focuses on integrating knowledge-based systems and neural networks for time-series analysis and forecasting.
\end{IEEEbiography}
\begin{IEEEbiography}[{\includegraphics[width=1in,height=1.25in,clip,keepaspectratio]{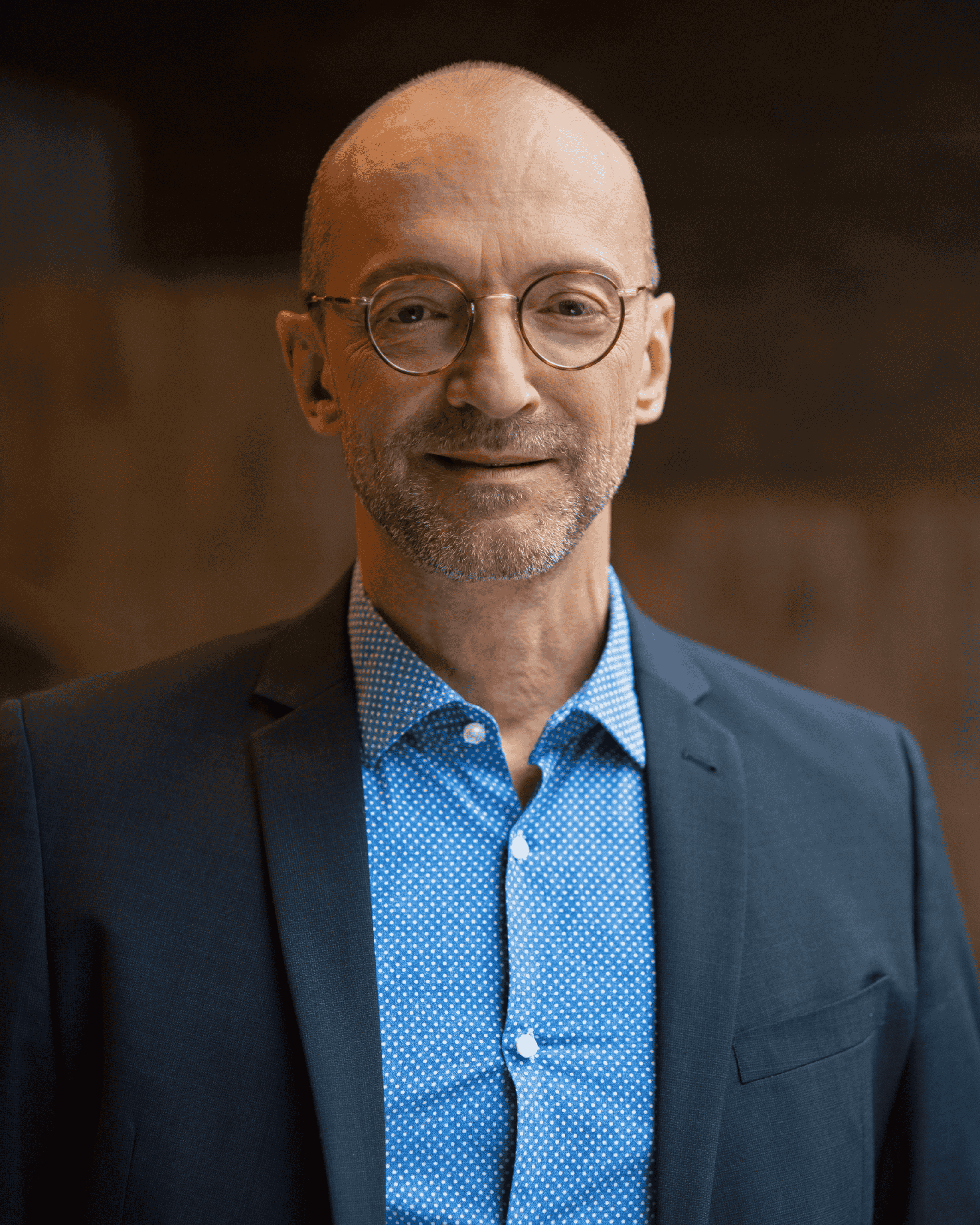}}]{Andreas Dengel} received the Diploma degree
in CS from TUK and the Ph.D. degree from the University of Stuttgart. He is currently the Scientific Director of DFKI GmbH, Kaiserslautern. In 1993, he became a Professor of computer science with TUK, where he holds the Chair of Knowledge-Based Systems. Since 2009, he has been appointed as a Professor (Kyakuin) with the Department of Computer Science and Information Systems, Osaka Prefecture University. He also worked at IBM, Siemens, and Xerox Parc. He is the co-editor of international computer science journals and has written or edited 12 books. He is the author of more than 300 peer-reviewed scientific publications and supervised more than 170 master's and Ph.D. theses. His main scientific research interests include pattern recognition, document understanding, information retrieval, multimedia mining, semantic technologies, and social media. He is a fellow of IAPR and received many prominent international awards. He is a member of several international advisory boards, has chaired major international conferences, and founded several successful start-up companies.
\end{IEEEbiography}
\begin{IEEEbiography}[{\includegraphics[width=1in,height=1.25in,clip,keepaspectratio]{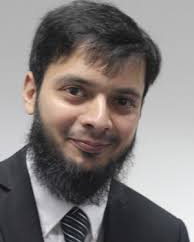}}]{Sheraz Ahmed} received the master's degree in computer science from Technische Universität Kaiserslautern, Germany, and the Ph.D. degree from German Research Center for Artificial Intelligence, Germany, under the supervision of Dr. H. C. A. Dengel and Dr. Habil M. Liwicki. His Ph.D. topic was generic methods for information segmentation in document images. From 2012 to 2013, he visited Osaka Prefecture University, Osaka, Japan, as a Research Fellow, supported by the Japanese Society for the Promotion of Science, and in 2014, he visited the University of Western Australia, Perth, Australia, as a Research Fellow, supported by the DAAD, Germany, and Go8, Australia. He is currently a Senior Researcher with German Research Center for Artificial Intelligence (DFKI GmbH), Kaiserslautern, where he is leading the area of time series analysis. Over the last few years, he has primarily worked on the development of various systems for information segmentation in document images. He has more than 30 publications on the said and related topics, including three journal articles and two book chapters. His research interests include document understanding, generic segmentation framework for documents, gesture recognition, pattern recognition, data
mining, anomaly detection, and natural language processing. He is a frequent reviewer of various journals and conferences, including Pattern Recognition Letters, Neural Computing and Applications, IJDAR, ICDAR, ICFHR, and DAS.
\end{IEEEbiography}
\begin{IEEEbiography}[{\includegraphics[width=1in,height=1.25in,clip,keepaspectratio]{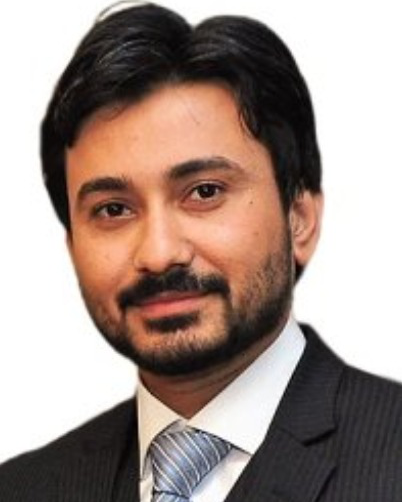}}]{Muhammad Naseer Bajwa} is a Postdoctoral Research Fellow at the School of Electrical Engineering and Computer Science (SEECS), National University of Sciences and Technology (NUST), Pakistan, and a Co-Principal Investigator at the Deep Learning Lab (DLL), National Centre of Artificial Intelligence (NCAI). He received his PhD from Germany, during which he conducted research at the German Research Centre for Artificial Intelligence (DFKI GmbH), focusing on the development of pragmatic, accurate, confident, and explainable computer-aided diagnosis frameworks. His research interests include trustworthy AI, uncertainty estimation, and explainable AI, with applications in healthcare, climate resilience, and sustainable systems. He has contributed to several internationally funded research projects in collaboration with partners in the UK and Europe and has served as a reviewer for multiple peer-reviewed journals and conferences, including MICCAI, ICDAR, and MIUA.
\end{IEEEbiography}
\begin{IEEEbiography}[{\includegraphics[width=1in,height=1.25in,clip,keepaspectratio]{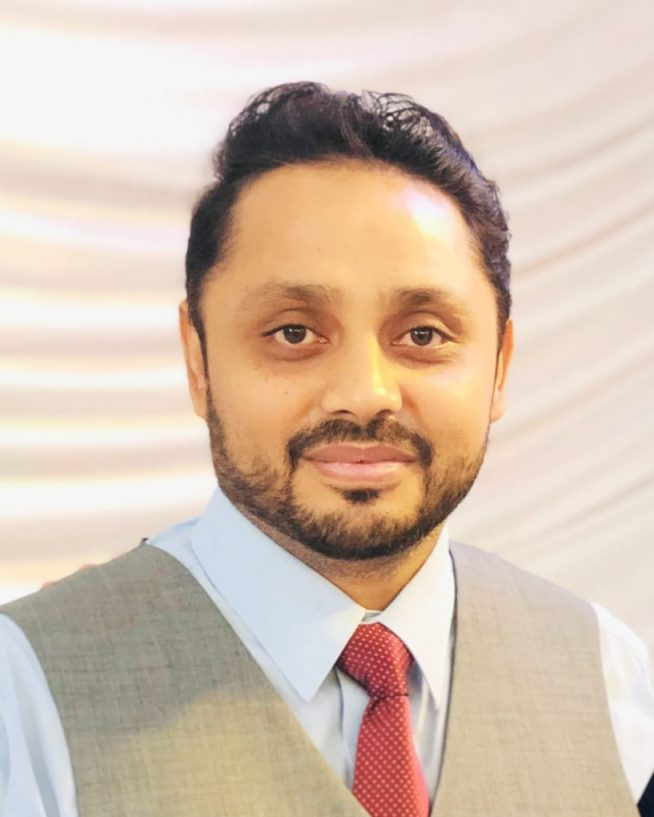}}]{Muhammad Imran Malik} received the master's and Ph.D. degrees in artificial intelligence from the University of Kaiserslautern, Germany, in 2011 and 2015, respectively. His Ph.D. topic was automated forensic handwriting analysis, on which he focused on both the perspectives of forensic handwriting examiners and pattern recognition researchers.
He was with the German Research Center for Artificial Intelligence GmbH (DFKI), Kaiserslautern. He is currently an Associate Professor and the Head of the Computer Science Department, School of Electrical Engineering and Computer Science, National University of Sciences and Technology, Islamabad, Pakistan. He has authored more than 80 publications, including high-ranked international conference papers, book chapters, and top-tier journal papers, which acquired more than 2500 citations on his Google Scholar profile. His research interests include machine learning and pattern recognition with a special emphasis on applications in image analysis and understanding.
\end{IEEEbiography}

\vfill

\end{document}